\documentclass[10pt,twocolumn,letterpaper]{article}

\usepackage{cvpr}
\usepackage{times}
\usepackage{epsfig}
\usepackage{graphicx}
\usepackage{amsmath}
\usepackage{amssymb}
\usepackage{enumitem}
\usepackage{mathrsfs}
\usepackage{booktabs}
\usepackage[caption=false]{subfig}
\usepackage{multicol}
\usepackage{gensymb}
\usepackage{mathtools, cases}
\usepackage{esvect}
\usepackage{color}

\usepackage[pagebackref=true,breaklinks=true,letterpaper=true,colorlinks,bookmarks=false]{hyperref}

\cvprfinalcopy 

\def\cvprPaperID{3305} 

\ifcvprfinal\fi
\begin{document}
	
	\title{Multi-person Implicit Reconstruction from a Single Image}

\author{Armin Mustafa$^{1}$  \hspace{.11\linewidth} Akin Caliskan$^{1}$  \hspace{.11\linewidth} Lourdes Agapito$^{2}$ \hspace{.11\linewidth} Adrian Hilton$^{1}$\\
	$^{1}$CVSSP, University of Surrey \hspace{.1\linewidth} $^{2}$Department of Computer Science, University College London
	}
	
	\maketitle
	\thispagestyle{empty}	
	\begin{abstract}
		We present a new end-to-end learning framework to obtain detailed and spatially coherent reconstructions of multiple people from a single image. Existing multi-person methods suffer from two main drawbacks: they are often model-based and therefore cannot capture accurate 3D models of people with loose clothing and hair; or they require manual intervention to resolve occlusions or interactions. Our method addresses both limitations by introducing the first end-to-end learning approach to perform model-free implicit reconstruction for realistic 3D capture of multiple clothed people in arbitrary poses (with occlusions) from a single image. Our network simultaneously estimates the 3D geometry of each person and their 6DOF spatial locations, to obtain a coherent multi-human reconstruction. In addition, we introduce a new synthetic dataset that depicts images with a varying number of inter-occluded humans and a variety of clothing and hair styles. We demonstrate robust, high-resolution reconstructions on images of multiple humans with complex  occlusions, loose clothing and a large variety of poses and scenes. Our quantitative evaluation on both synthetic and real world datasets demonstrates state-of-the-art performance with significant improvements in the accuracy and completeness of the reconstructions over competing approaches.
	\end{abstract}
	
	\section{Introduction}
\begin{figure}
		\begin{center}
			\includegraphics[width=0.49\textwidth]{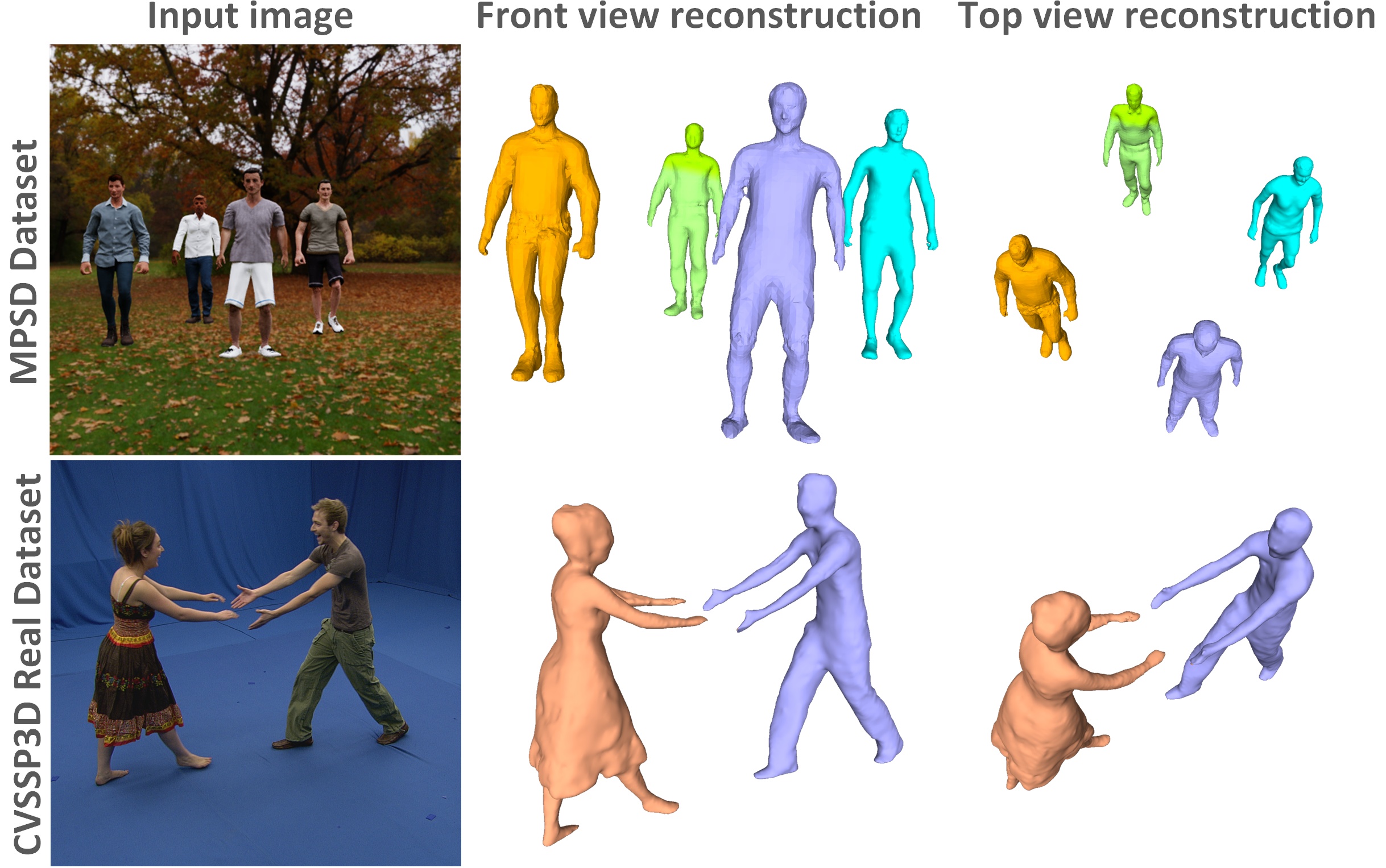}
			\caption{Proposed model-free multi-person spatially coherent implicit reconstruction from a single image with 4 and 2 people on synthetic MPSD and real CVSSP3D dataset.}
			\label{fig:motivation}
			\vspace{-0.2cm}
		\end{center}
	\end{figure}

	Multi-person human reconstruction from a single image finds application in surveillance; film and entertainment including movie production; generating AR/VR content for complex scenes; and sports broadcast. Reconstruction from a single camera is more practical and lower-cost compared to multi-view, as it does not require a complex setup.
	Immense progress has been made in estimating 3D human pose and shape from a single image or monocular video in the last five years \cite{alldieck2019tex2shape,zheng2019deephuman,Saito_2019_ICCV,GabeurFMSR2019,Guler_2019_CVPR}. Methods can be classified as model-based or model-free. Model-based methods use a parametric human body shape model such as SMPL to reconstruct people from a single image \cite{bogo2016smpl,Lassner2017,hmrKanazawa17} including methods that estimate SMPL with clothing top\cite{Ma_2020_CVPR}. Model-free methods give a more realistic reconstruction of people with loose clothing and hair details \cite{alldieck2019tex2shape,zheng2019deephuman,Saito_2019_ICCV,GabeurFMSR2019}. However all  existing methods require an image of a single fully visible person without occlusions to allow reconstruction. Recently, model-based approaches have been introduced that can reconstruct multiple humans in a scene \cite{Guler_2019_CVPR,jiang2020coherent,Zhang2020Perceiving3H} using SMPL, so cannot capture clothing details. In addition,   \cite{Zhang2020Perceiving3H} requires manual intervention to mark interaction regions on 3D surfaces to handle inter-person/object occlusions. This paper introduces the first model-free end-to-end approach that reconstructs multiple clothed people from a single image of a crowded scene with inter-person occlusions (see Table \ref{t_litsurvey}). Our proposed approach produces a spatially coherent implicit reconstruction of each person together with their 6DOF spatial locations and orientations in the observed scene without any manual intervention, and can handle complex poses, clothing and partial occlusion from a single image, as shown in Fig. \ref{fig:motivation}.

	We introduce the first multiple people synthetic dataset and benchmark (MPSD) with realistic image-3D model pairs. MPSD ranges from $2-10$ people per image in a wide variety of clothing, hairstyles and poses with detailed surface geometry and appearance rendered with diverse indoor and outdoor natural backgrounds and realistic scene illumination. This dataset provides the first quantitative benchmark for multi-person single image reconstruction. We propose an end-to-end method trained on the MPSD dataset that estimates 3D reconstructions of each person and their 6DOF location/orientation by exploiting single image depth and instance segmentation. Example results from the proposed single image multi-human spatially coherent implicit reconstruction are shown in Fig. \ref{fig:motivation}. Our contributions are:
	\begin{itemize}[topsep=0pt,partopsep=0pt,itemsep=0pt,parsep=0pt] 
		\item The first approach for model-free reconstruction of multiple people from a single image with accurate spatial arrangement.
		\item An end-to-end framework using cascaded multitask networks for simultaneous implicit 3D reconstruction and 6DOF location/orientation estimation exploiting depth and instance segmentation information.
		\item A multiple person synthetic image/3D dataset of complex multi-person scenes with inter-person occlusions, realistic clothing, hair, poses, scenes and illumination.
		\item A method that exploits the advantages of volumetric and implicit 3D shape representations for detailed reconstructions of clothed people from a single image.
	\end{itemize}
	%
	\begin{table}
		\setlength{\tabcolsep}{2pt}
		\scalebox{0.9}{
			\begin{tabular}{l|c|c|c|c|c}
				& \textbf{Model-free} & \textbf{Multi-human} & \textbf{Coherent} & \textbf{Occ.} & \textbf{RGB}\\ \hline
				\cite{bogo2016smpl,Lassner2017,hmrKanazawa17,kolotouros2019spin} & \textcolor{red}{$\times$}  & \textcolor{red}{$\times$} & \textcolor{red}{$\times$} & \textcolor{red}{$\times$} & \checkmark \\ \hline
				\cite{alldieck2019tex2shape,zheng2019deephuman,Saito_2019_ICCV,GabeurFMSR2019} & \checkmark & \textcolor{red}{$\times$} & \textcolor{red}{$\times$} &  \textcolor{red}{$\times$} & \checkmark \\ \hline
				\cite{jiang2020coherent,Zanfir2018Monocular3P,NIPS2018_8061} & \textcolor{red}{$\times$} & \checkmark & \checkmark & \checkmark & \checkmark \\ \hline
				\cite{bhatnagar2020ipnet,Chibane_2020_CVPR} & \checkmark & \textcolor{red}{$\times$} &\textcolor{red}{$\times$} & \checkmark & \textcolor{red}{$\times$} \\ \hline
				Holopose \cite{Guler_2019_CVPR} & \textcolor{red}{$\times$}  & \checkmark & \textcolor{red}{$\times$} & \checkmark & \checkmark \\ \hline
				PHOSA \cite{Zhang2020Perceiving3H} & \textcolor{red}{$\times$} & \checkmark & \checkmark & \checkmark & \checkmark \\ \hline
				\textbf{Proposed} & \checkmark & \checkmark & \checkmark  & \checkmark & \checkmark
		\end{tabular}}
		\caption{Comparison of our method with existing 3D shape estimation methods. Occ - Occlusions and Image - Single image}
		\label{t_litsurvey}
	\end{table}
	\begin{figure*}[t]
		\begin{center}
			\includegraphics[width=0.98\textwidth]{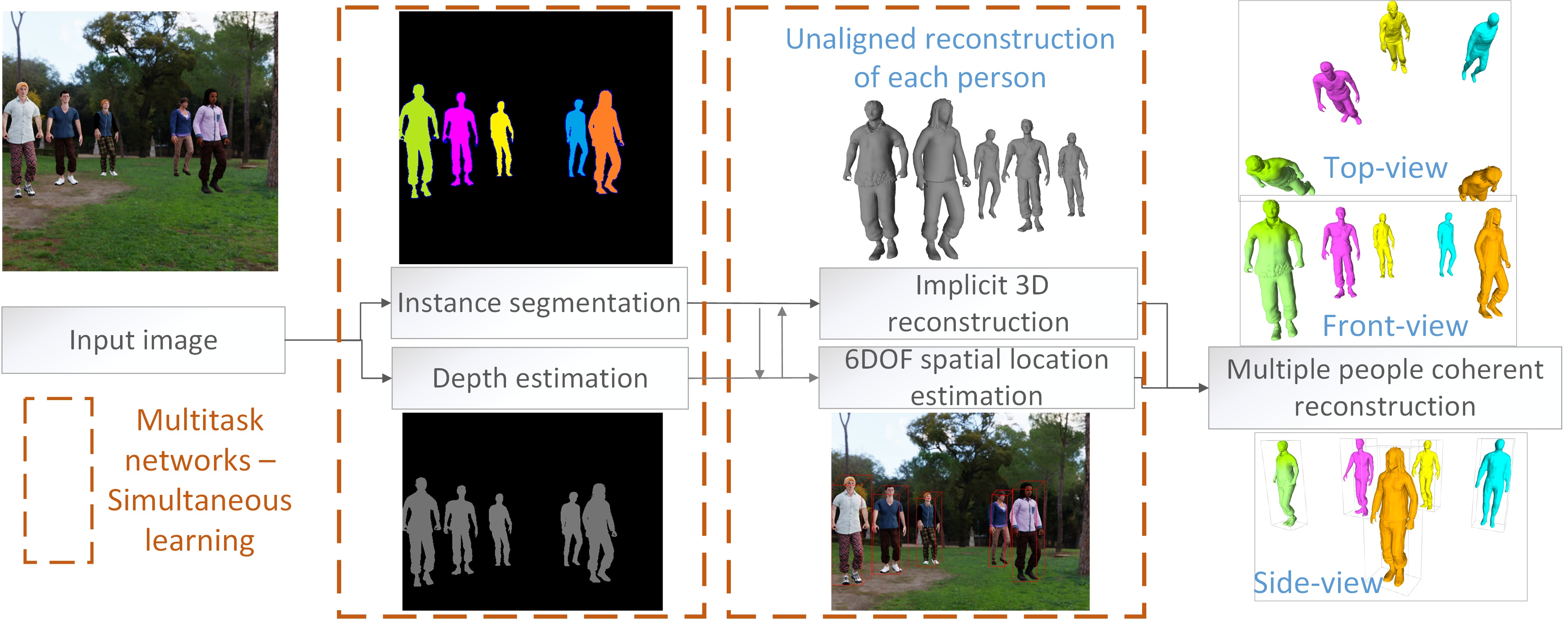}
			\caption{ Proposed end-to-end model-free spatially coherent multi-person implicit reconstruction from a single image. The first multitask network estimates segmentation and depth from a single image. The segmentation and depth map is used in the second multitask network to estimate per-instance implicit surface reconstruction of each person together with their 6DOF location and orientation.}
			\label{fig:algo}
			\vspace{-0.8cm}
		\end{center}
	\end{figure*}
	\section{Related Work}
	This section reviews work in the area of 3D human shape estimation from a single image for a single person and multiple people in the scene, an overview is shown in Table \ref{t_litsurvey}.
	\subsection{Single Person 3D Shape Estimation}
	Early techniques for single image 3D shape estimation were model-based \cite{GuanICCV2009,bogo2016smpl,hmrKanazawa17,kolotouros2019spin} with most approaches using the SMPL \cite{SMPL}. Methods fit the SMPL model to a single person in an image by using 2D joints \cite{bogo2016smpl}, silhouettes \cite{Lassner2017}, and 3D joints \cite{hmrKanazawa17}. 2D Joints and silhouettes were used for 3D shape estimation in \cite{Pavlakos2018} with the SMPL-X model \cite{SMPL-X:2019} and \cite{Ma_2020_CVPR,Yu2019SimulCapS,bhatnagar2019mgn} estimate tight fitting clothing on top of the SMPL model. However human shape estimated from these approaches does not represent the actual shape of humans with loose clothing, hair details and large deformations.
	
	Monocular model-free approaches estimate detailed shape of clothed people \cite{varol18bodynet,zheng2019deephuman,alldieck2019tex2shape,GabeurFMSR2019,Saito_2019_ICCV,Dong_2019_ICCV,Chibane_2020_CVPR} without a parametric model. \cite{varol18bodynet} uses a voxel representation, \cite{zheng2019deephuman} uses normals to obtain a discretized volumetric representation, \cite{GabeurFMSR2019} uses front and back depth maps and \cite{alldieck2019tex2shape} uses the SMPL model as an initialization to predict model-free 3D human shape. \cite{Yu2019SimulCapS} estimates 3D from a RGBD camera and \cite{natsume_siclope2019} uses multi-view silhouettes to obtain 3D shape from a single image. PiFU regresses an implicit function to determine the occupancy for any given 3D location \cite{Saito_2019_ICCV}. This was superseded by PiFUHD for highly detailed 3D reconstruction \cite{saito2020pifuhd}. PiFU and PiFUHD do not work well for arbitrary poses. This limitation was partially addressed in \cite{huang2020arch} by using a semantic deformation field. \cite{bhatnagar2020ipnet,Chibane_2020_CVPR} estimates 3D human shape as an implicit function from a sparse point cloud instead of an RGB image, exploiting the SMPL model. However all these methods reconstruct a single person with a limited range of pose and require full visibility without occlusions. \cite{caliskan2020multiview} exploits multi-views during training to estimate 3D human shape from a single image in a wide variety of poses. This method uses a voxel representation which results in a quantized output mesh with low resolution and less accurate reconstruction.
	
	In this paper, we introduce a method that exploits the advantages of both volumetric voxel and implicit representations by obtaining an initial voxelized mesh of a person which is refined using an implicit function. The proposed method is trained on multiple views \cite{caliskan2020multiview} to handle partially occluded people in a wide variety of poses.
	\subsection{Multiple Person 3D Shape Estimation}
	There are a limited number of methods that predict 3D shape of multiple humans from a single image. \cite{Zanfir2018Monocular3P,NIPS2018_8061} were the first methods to perform multi-human spatially coherent reconstruction from a single image exploiting SMPL by using multiple scene constraints to optimize 3D shape \cite{Zanfir2018Monocular3P} and a feed-forward network to estimate pose and shape for multiple people \cite{NIPS2018_8061}. Jiang et al \cite{jiang2020coherent} performed more accurate and robust SMPL-based multi-human reconstruction by directly regressing the SMLP parameters from pixels. 
	Holopose \cite{Guler_2019_CVPR} reconstructs multiple people from a single image using Densepose \cite{guler2018densepose}, but with a single scale and no spatial layout.
	Recently \cite{Zhang2020Perceiving3H} introduced an optimization framework for multi-human and object reconstruction using collision and depth information but it required manual intervention to estimate heights of each class and mark the interaction regions on 3D mesh. However all these methods require the SMPL model and suffer from the same limitations as model-based approaches for single humans. 
	
	Our paper introduces the first end-to-end model-free method to implicitly reconstruct multiple humans with loose clothing and hair details in a wide variety of poses with inter-person occlusions from a single crowded image, which is a non-trivial task and an unsolved problem in the literature. The proposed end-to-end method has two multitask networks: the first  network estimates instance segmentation and depth from the input image; which is exploited in the second  network that simultaneously estimates the 3D shape and 6DOF spatial location and orientation of each person to achieve a spatially coherent implicit 3D reconstruction from a single image.
	\section{Methodology}
	\subsection{Overview:}
	\label{sec:overview}
	Our proposed learning-based method performs multi-person spatially coherent reconstruction without any manual intervention by first estimating instance segmentation and depth using existing multitask encoder-decoder network \cite{kendall2017multi} from a single image, followed by the second multitask network that estimates implicit 3D reconstruction and 6DOF spatial location of each person simultaneously exploiting the depth and instance segmentation, as shown in Fig. \ref{fig:algo}. The proposed end-to-end network is trained on a novel \textit{MPSD} dataset introduced in this paper.
	
\begin{figure*}
		\begin{center}
			\includegraphics[width=0.95\textwidth]{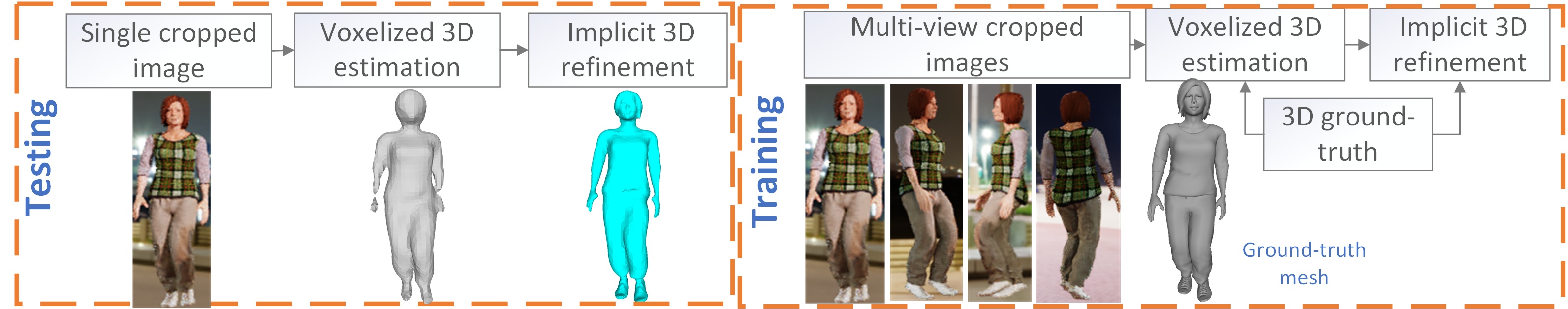}
			\caption{Implicit 3D reconstruction - testing and training framework.}
			\label{fig:implicit}
			\vspace{-0.75cm}
		\end{center}
	\end{figure*}

	\noindent
	\textbf{Implicit 3D reconstruction, Sec. \ref{sec:3d}, Fig. \ref{fig:implicit}:} For each human instance an implicit 3D surface is estimated using an intermediate volumetric voxel-based representation to allow reconstructions from a wide range of poses. The intermediate representation is obtained using multi-view voxel based 3D shape estimation, inspired by \cite{caliskan2020multiview}. \cite{caliskan2020multiview} is trained on multiple views for a more accurate reconstruction from a single image. A mutli-view occlusion silhouette loss is added to improve the 3D output \cite{Zhang2020Perceiving3H}. However this gives a quantized voxel output with low resolution. This is addressed in this paper through an additional network that performs implicit function based refinement on the voxel output giving a high-resolution and more complete 3D shape of clothed people exploiting features from voxels, image and depth using hybrid representation learning. The multi-view training in the proposed approach enables us to handle inter-person occlusions and partial visibility in crowded images.
	
	\noindent
	\textbf{6DOF spatial location/orientation estimation, Sec. \ref{sec:pose}, Fig. \ref{fig:pose}:} The per-person 3D reconstructions from the implicit refinement are unaligned in different coordinate systems leading to spatially incoherent output and incorrect world coordinates. To create a spatially coherent 3D reconstruction from a single image, 6DOF spatial location and orientation is estimated for each person. The instance segmentation and depth is exploited in 6DOF location estimation, inspired by a recent 6DOF pose method \cite{Wang2019DenseFusion6O}. However \cite{Wang2019DenseFusion6O} uses only local depth features, which gives limited performance. We add both local and global depth features along with an ordinal depth loss \cite{jiang2020coherent} to the network, to improve the location/orientation estimation in crowded scenes.
	
	The proposed network is trained on a new Multiple People Synthetic Dataset (\textit{MPSD}) with image/3D pairs of varying number of people in the images, 3D ground-truth, instance segmentation and depth maps. This trained network simultaneously predicts 3D shapes and spatial layouts of each person for multi-human 3D reconstruction from a single image with correct relative spatial arrangement. 
	%
	\begin{figure}[h]
		\begin{center}
			\includegraphics[width=0.49\textwidth]{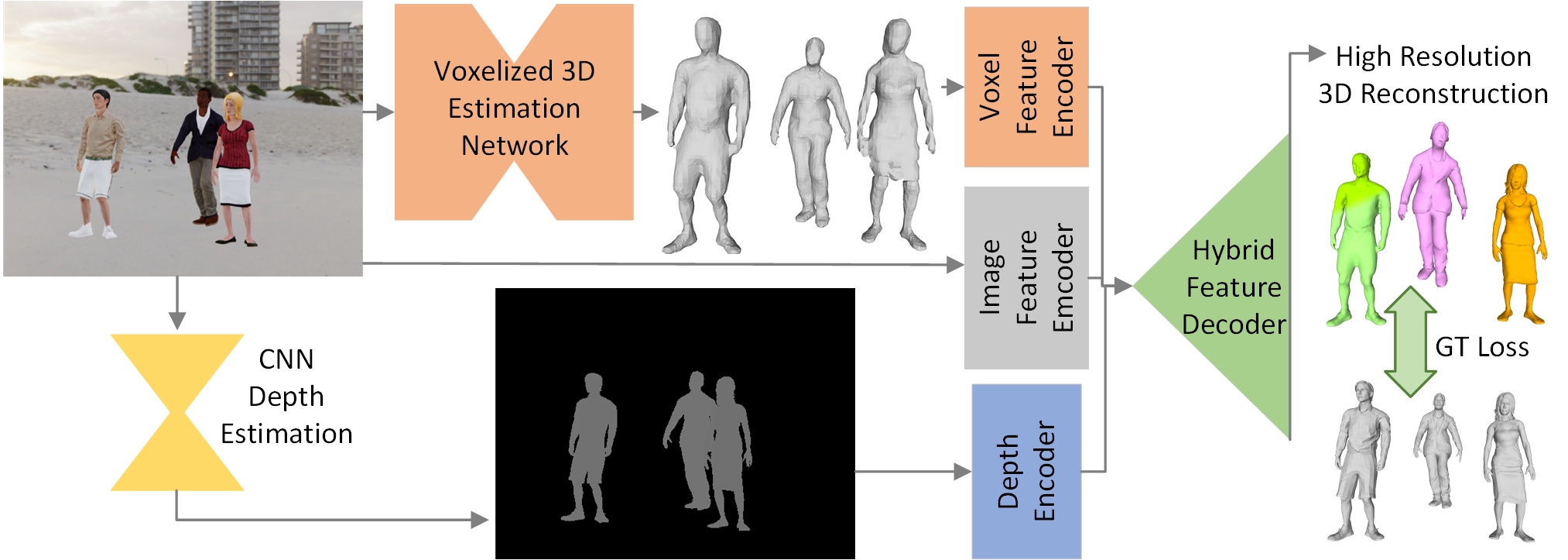}
			\caption{Implicit 3D refinement framework.}
			\label{fig:ImplicitR}
			\vspace{-0.2cm}
		\end{center}
	\end{figure}
	\subsection{Implicit 3D Reconstruction}
	\label{sec:3d}
	Existing monocular model-free 3D shape estimation methods either use a volumetric \cite{varol18bodynet} or implicit\cite{Saito_2019_ICCV} representation. The volumetric methods work for a wide variety of poses, clothing and hair, however the output 3D is low resolution which lacks surface details. Implicit representation gives a highly detailed surface, but is limited to restricted poses and clothing \cite{huang2020arch,caliskan2020multiview}. In this paper we combine volumetric and implicit representations to benefit from advantages of both, as shown in Fig. \ref{fig:implicit}. In addition our approach handles partial occlusion which none of the previous implicit or volumetric approaches to model-free single human reconstruction allow. The first stage gives an intermediate 3D volumetric voxel representation \cite{caliskan2020multiview} which handles variation in human poses, clothing and hair and the second stage refines the surface using implicit representation for a more complete high-quality detailed surface reconstruction handling inter-person occlusions. The two stages are:
	
	\noindent
	\textbf{Voxelized 3D estimation:} Intermediate voxelized 3D is obtained using the method \cite{caliskan2020multiview}, which outperforms implicit reconstruction methods \cite{Saito_2019_ICCV} in completeness and accuracy of human shape of visible and occluded parts of humans in a wide variety of poses, clothing and hairstyles because of multi-view training. Similarly the proposed method is trained on multiple views for voxelized 3D estimation of each person, as shown in Fig. \ref{fig:implicit}, using the loss function: $L = L_{3D} + \alpha L_{M} + \beta L_{OS}$. We have added occlusion silhouette loss ($L_{OS}$) \cite{Zhang2020Perceiving3H} in addition to the 3D ground-truth loss ($L_{3D}$) and multi-view consistency loss ($L_{M}$) \cite{caliskan2020multiview}.
	\vspace{-0.5cm}
	\begin{multline} \nonumber
		{L}_{3D} = \sum_{p\in V}\sum_{n=1}^{N} \lambda O_{p}^{n} \log\widehat{O}_{p}^{n} + (1-\lambda)(1-O_{p}^{n}) (1-\log\widehat{O}_{p}^{n})\\
		{L}_{M} = \sum_{p\in V}\sum_{\substack{n=l=1\\  l \neq n }}^{N} \left \|\widehat{O}_{p}^{n} -  \widehat{O}_{p}^{l} \right \|_{2} ;
		L_{OS} = \sum_{n=1}^{N} \sum_{i\in I} S_{n}^{i} - m\cdot\hat{S}_{n}^{i}
	\end{multline}
	where $N$ is the number of views, $O_{p}$ is the occupancy value of 3D point $p$ in the voxel grid $V$, $\widehat{O}_{p}$ is the predicted value, $S$ is the ground-truth silhouette $\hat{S}$ is the rendered silhouette, $m$ is the visibility indicator and $\alpha,\lambda,\beta$ are constants chosen experimentally, defined in Sec. \ref{sec:implement}. The multi-view occlusion silhouette loss allows the proposed method to learn 3D shape robust to camera view changes and self-occlusion (Sec. \ref{sec:ablation} for more details). 
	We choose $N=4$ different views $90 \degree$ apart for training for a balance between memory use and performance. Multiple people images often suffer from occlusions, multi-view training enables us to address inter-person occlusions by reliably reconstructing the unseen parts of a human using the volumetric representation surface reconstruction. However the surface details are limited due to the relative low resolution of the discrete 3D voxel output. To address this resolution limitation a novel implicit 3D refinement stage is proposed. 

	\noindent
	\textbf{Implicit 3D refinement:} 
	To obtain high-resolution surface detail from the low resolution 3D voxel reconstruction, an implicit refinement is proposed as shown in Fig. \ref{fig:implicit}. This refinement uses features from the voxelized output, the input image and the predicted depth map, which are fed into the proposed decoder to implicitly compute the occupancy value of an arbitrary 3D point.
	The proposed decoder takes three inputs: first input is the \textit{voxel feature} extracted from the voxelized 3D of the first stage using the multi-dimensional voxel encoder \cite{Chibane_2020_CVPR} and the sampled 3D point; second input is the pixel-wise \textit{image feature} extracted from the input RGB image using the hourglass network \cite{Saito_2019_ICCV}; and third input is the depth of the corresponding 3D point with respect to the camera view. In the decoding stage, the occupancy value of the 3D point is predicted from the hybrid feature representations. L1 loss is applied between the predicted occupancy ($\widehat{O}$) and the ground-truth occupancy ($O$) defined as: $L_{GT} = \sum_{p\in \mathscr{M}}\left | O_{p} - \widehat{O}_{p} \right |_{1}$, where $p$ is a 3D point on mesh $\mathscr{M}$. The implicit function for the predicted occupancy is defined as: $\widehat{O} = f(\phi, \varphi, d) \in \left [ 0,1 \right ]$, where $\phi$ are image features, $\varphi$ are point-wise voxel features and $d$ are depth features. The 3D occupancy values are used to create meshes for each person using Marching cubes \cite{Lorensen87marchingcubes}.The details of the decoder and ablation on the hybrid features are given in Sec. \ref{sec:experiment}. The results of 3D reconstruction before and after \textit{voxel refinement} are illustrated in Fig. \ref{fig:implicitablation}.
	\subsection{Multiple People Synthetic Dataset (MPSD)}
	\label{sec:dataset}
	Existing datasets for training 3D shape estimation methods either have a single person (Surreal\cite{varol18bodynet}, THuman \cite{zheng2019deephuman}, 3D-Humans \cite{GabeurFMSR2019}, 3DPeople\cite{pumarola20193dpeople}, 3DVH\cite{caliskan2020multiview}) or do not have ground-truth 3D models for multiple people \cite{MultiHumanflow,XNect_SIGGRAPH2020}.
	Training the proposed end-to-end multi-person spatially coherent reconstruction network requires ground-truth 3D human models and their respective 6DOF spatial location/orientation for multiple people. Hence we introduce the first and the largest realistic Multiple People Synthetic Dataset \textit{(MPSD)} generated with synthetic humans in a wide variety of clothing, poses and hair styles in arbitrary positions simulated against different realistic backgrounds. The dataset improves the generalisation of multiple human reconstruction in different poses, positions, clothing etc.
	\begin{figure}
		\begin{center}
			\includegraphics[width=0.49\textwidth]{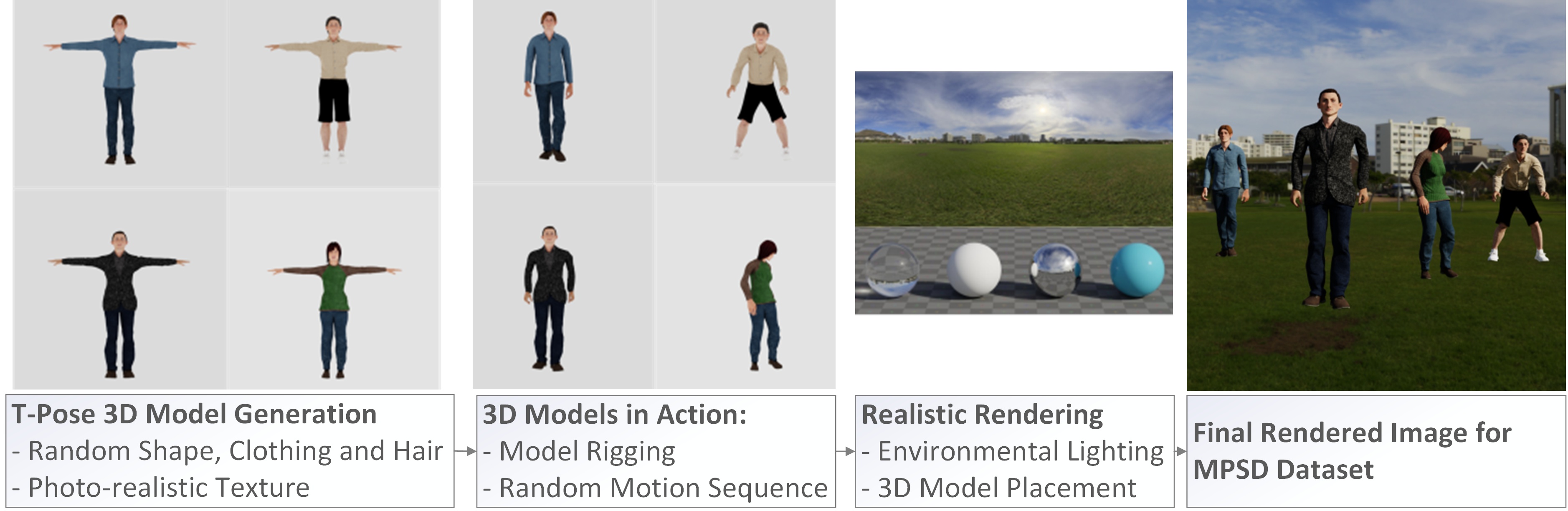}
			\caption{The MPSD dataset generation framework}
			\label{fig:mv3dvhDataset}
			\vspace{-0.2cm}
		\end{center}
	\end{figure}
	\begin{figure}
		\begin{center}
			\includegraphics[width=0.49\textwidth]{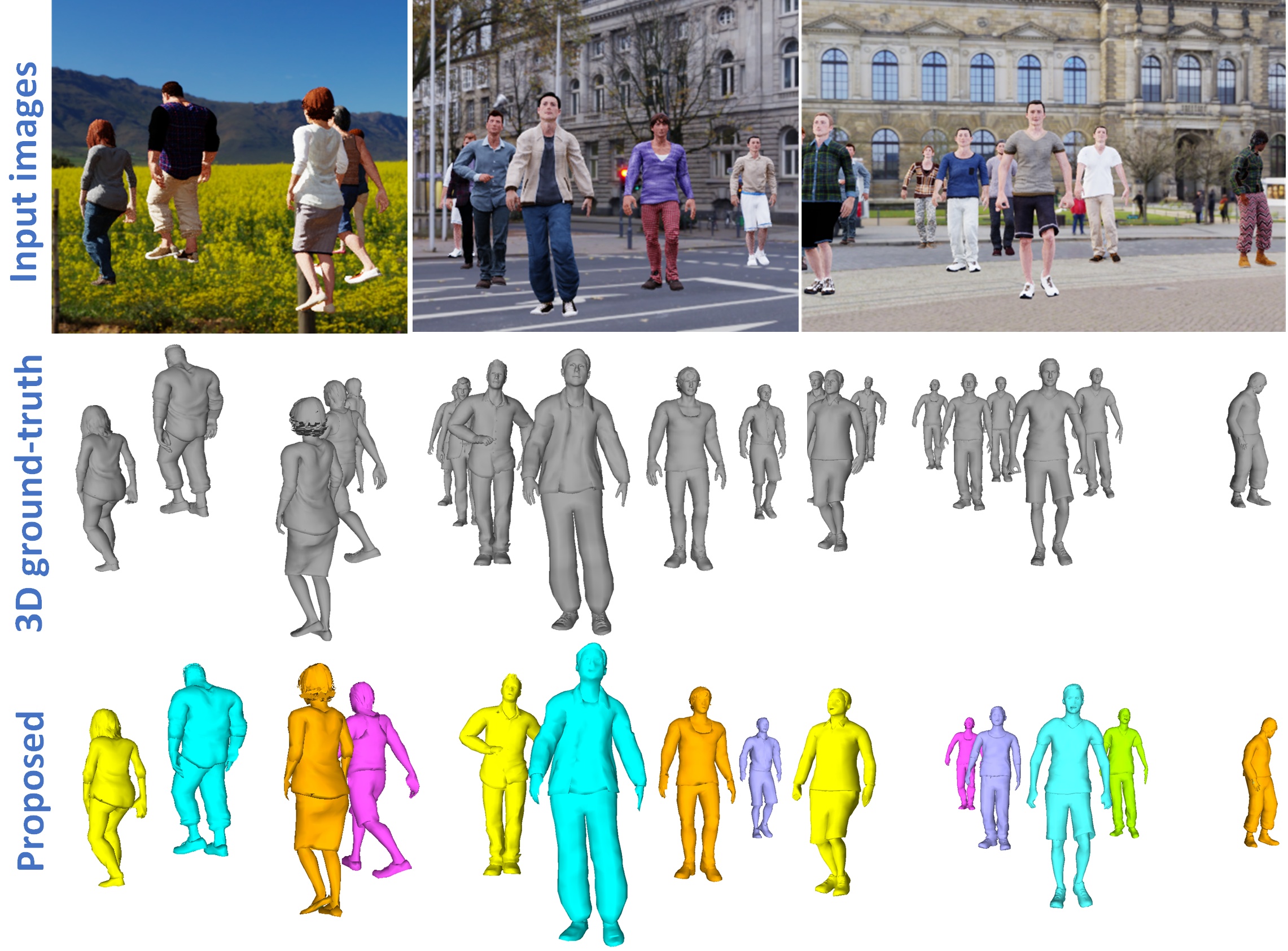}
			\vspace{-0.4cm}
			\caption{Examples of multi-person images from the MPSD dataset. The proposed method is able to reconstruct partially occluded people in the scene but not heavily occluded.}
			\label{fig:mpsd}
			\vspace{-0.2cm}
		\end{center}
	\end{figure}
	\begin{figure}[t]
		\begin{center}
			\includegraphics[width=0.49\textwidth]{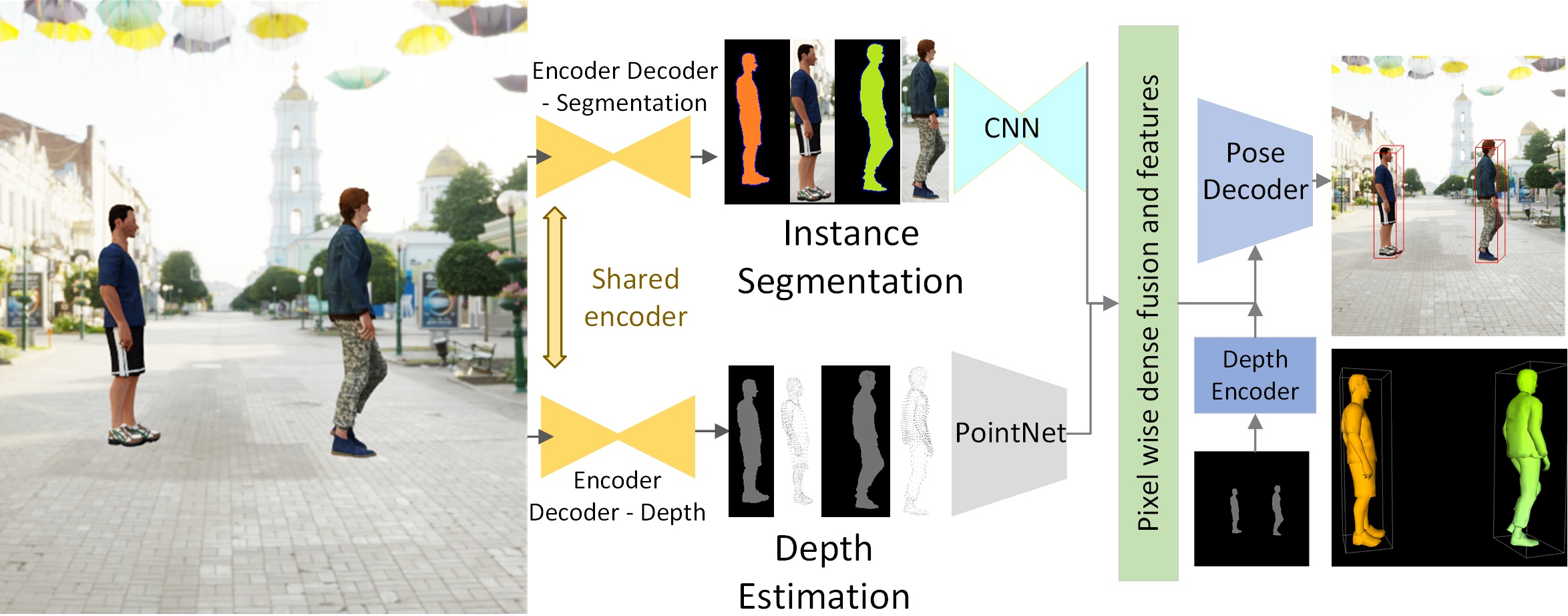}
			\caption{6DOF spatial location/orientation estimation framework}
			\label{fig:pose}
			\vspace{-0.2cm}
		\end{center}
	\end{figure}
	\begin{figure}[t]
		\begin{center}
			\includegraphics[width=0.49\textwidth]{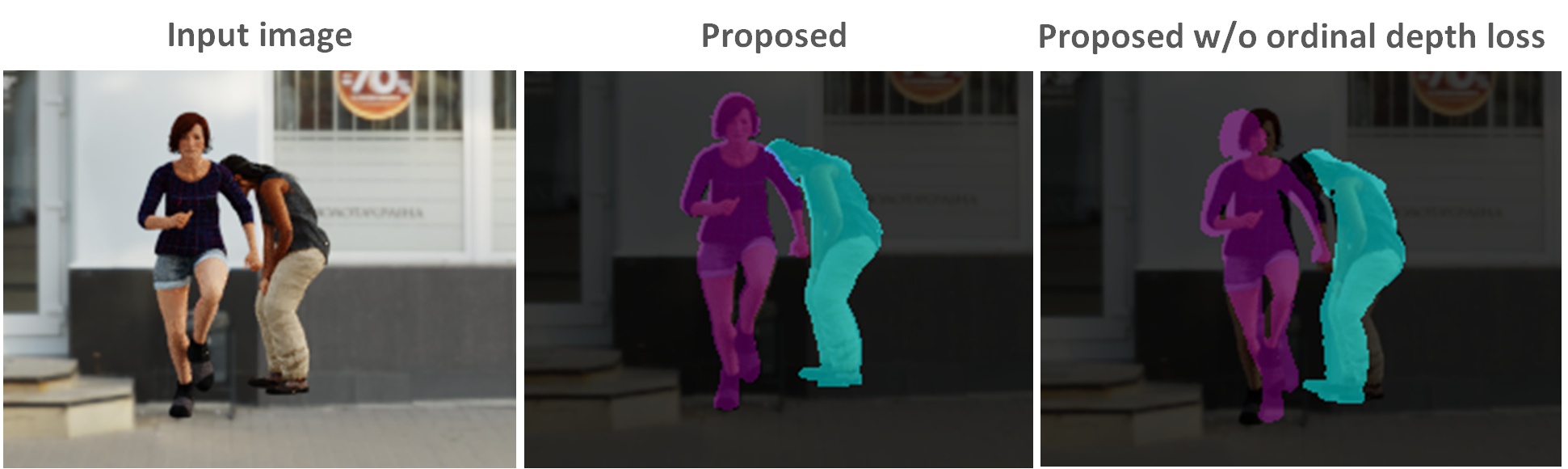}
			\caption{Comparison of location estimated from the proposed method without the global depth features and ordinal depth loss.}
			\label{fig:posecompare}
			\vspace{-0.2cm}
		\end{center}
	\end{figure}
	
	The \textit{MPSD} dataset is generated in three steps, as seen in Fig. \ref{fig:mv3dvhDataset}: clothed 3D human model generation; motion sequence application on these models; and multi-view realistic rendering of the models with random placement without intersections. $400$ male and female 3D human models with a wide variation in hair, clothing, pose and random positions are generated \cite{adobefuse} for various motion sequences \cite{adobemixamo} for more complete and accurate 3D shape estimation. 
	%
	This dataset consists of varying number of people $\{ 2,3, \dots ,9,10 \}$ at random positions in the scene, as seen in Fig. \ref{fig:mpsd} along with the results from the proposed method. The \textit{MPSD} dataset contains $450k$ \textit{image} - \textit{3D models} pairs which are used for single-image multi-human reconstruction. The scene is rendered into 16 camera views with a $512 \times 512$ image resolution at each time instant. The \textit{MPSD} dataset provides 3D ground-truth human models, RGB images, depth, instance segmentation and 6DOF spatial locations, which is used to train end-to-and proposed network (Sec. \ref{sec:overview}).
	%
	The \textit{MPSD} dataset will be released to support research and benchmarking. We will provide RGB images, 6DOF spatial locations, depth maps, instance segmentation and a framework for users to reproduce 3D models, in compliance with the Adobe FUSE licensing terms for release.
	%
	\subsection{6DOF Spatial Location/Orientation Estimation}
	\label{sec:pose}
	6DOF spatial location/orientations are estimated for each 3D instance, as shown in Fig. \ref{fig:pose} to obtain a spatially coherent reconstruction consistent from different views. Previous approaches for multi-human spatially coherent reconstruction either incorporate coherency constraints within the SMPL model estimation \cite{jiang2020coherent} or use an optimization framework to estimate the 6DOF pose and scale \cite{Zhang2020Perceiving3H}. In this paper we estimate 6DOF location and orientation for spatially coherent reconstruction as the proposed implicit reconstruction gives shape at the same scale as the input image, removing the requirement of estimating scale.
	
	Out of the many existing 6DOF object pose approaches \cite{Wang2019DenseFusion6O,Runz_2020_CVPR,xiang2017posecnn,tian2020robust}, we choose \cite{Wang2019DenseFusion6O} as our baseline method as it uses segmentation and RGBD information within the network instead of other methods that either use only RGB image and depth separately or use costly post-processing steps, limiting their performances in crowded scenes. We use CNN and PointNet networks from \cite{Wang2019DenseFusion6O} to extract features from segmentation and depth and combine them using a dense fusion network to extract pixel-wise dense feature embedding to estimate pose \cite{Wang2019DenseFusion6O}, as shown in Fig. \ref{fig:pose}. Feature embedding networks and pose decoders are used to estimate the 6DOF location/orientation. In addition to the local depth features \cite{Wang2019DenseFusion6O} we also use global depth features in the pose decoder to improve the location estimation, as seen in Fig. \ref{fig:posecompare}.
	The network is trained using a combination of dense pose loss ($L_{DP}$) \cite{Wang2019DenseFusion6O} and ordinal depth loss ($L_{OD}$) \cite{jiang2020coherent}, $L_{Pose} = L_{DP} + \gamma L_{OD}$ defined as:\\
	$L_{DP} = \frac{1}{U} \sum_{i\in U} \left ( \frac{c_{i}}{Q}\sum_{j\in Q} \left \| (Rx_{j} + t) + (\hat{R_{i}}x_{j} + \hat{t_{i}}) \right \| \right ) \\ - \left ( w \log c_{i} \right )$ ; \\
	$L_{OD} = \sum_{i \in \mathcal{S}} \log \left ( 1 + \exp(D_{y(i)}(i) + D_{\hat{y}(i)}(i)) \right )$\\
	where $R,t$ are ground-truth pose and $\hat R,\hat t$ are predicted poses, $U$ is randomly sampled dense-pixel features, $Q$ is randomly selected 3D points, $c$ is confidence score for each prediction, $w$ is weight selected empirically and $D()$ is the depth value. The ordinal depth loss is applied on the entire depth image which is encoded in the form of features and is input to the pose decoder, as seen in the Fig. \ref{fig:pose}. This allows for a more accurate pose estimation in the case of humans in close proximity, as seen in Fig. \ref{fig:posecompare}. The proposed method with global depth encoded features and ordinal depth loss has been evaluated to perform the best.
	\begin{figure}[t]
		\begin{center}
			\includegraphics[width=0.48\textwidth]{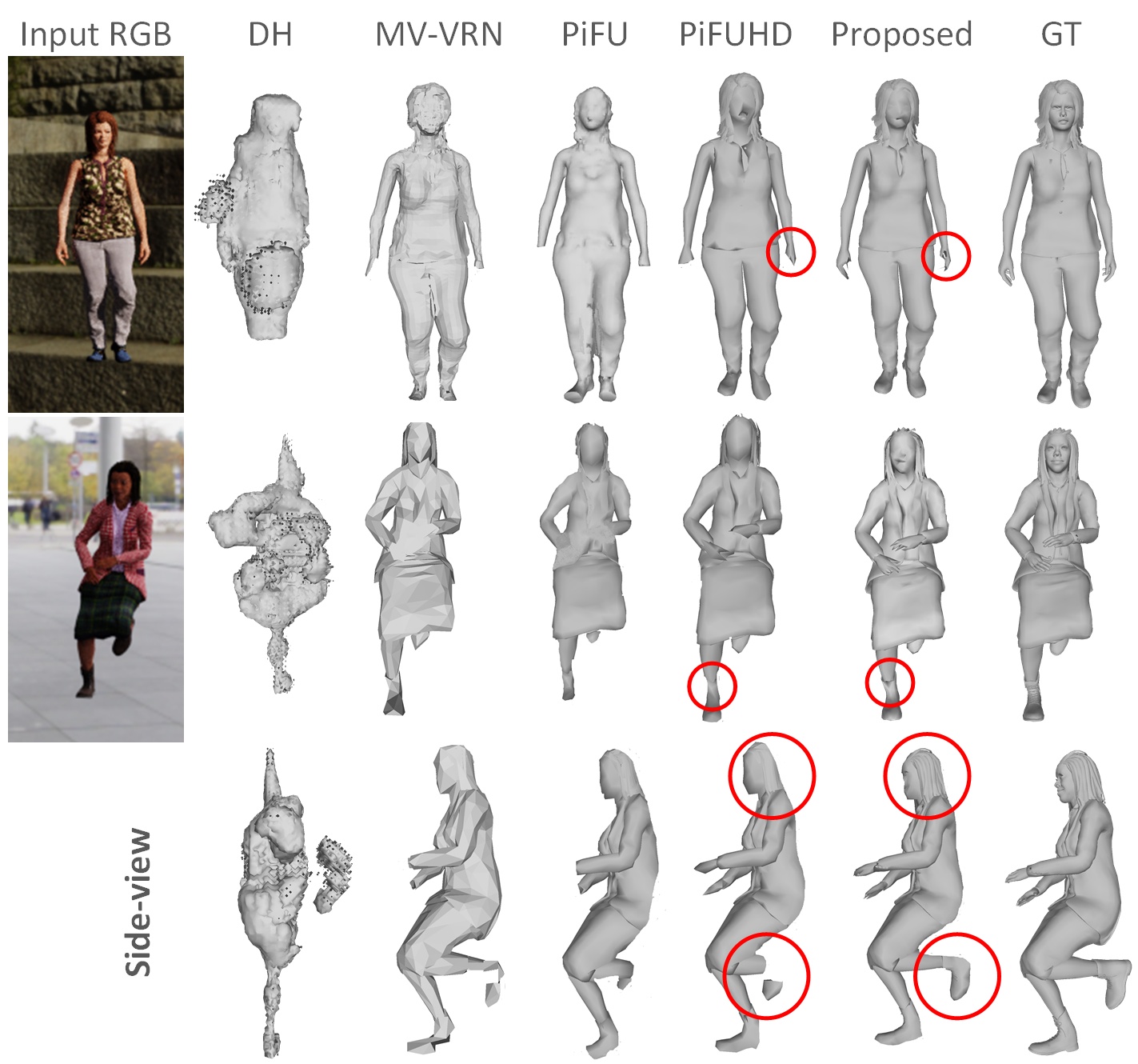}
			\caption{Comparison of proposed method against state-of-the-art single image shape estimation methods - GT is ground-truth and DH is DeepHuman. Differences are highlighted in red circle.}
			\label{fig:comparesingle}
			\vspace{-0.3cm}
		\end{center}
	\end{figure}
	\section{Evaluations and Results}
	\label{sec:experiment}
	An extensive experimental evaluation is presented on multiple people \cite{Zhang2020Perceiving3H,Kocabas2020VIBEVI} and single person \cite{Saito_2019_ICCV,saito2020pifuhd,caliskan2020multiview,zheng2019deephuman} 3D shape estimation against state-of-the-art methods on the proposed MPSD dataset and publically available single \cite{Vlasic_dynamicshape,zheng2019deephuman} and multi person \cite{Mustafa19,MustafaIJCV2019,UnstructuredVBR10} datasets. We were unable to compare with \cite{jiang2020coherent} because of unavailability of code and the datasets they used do not have 3D shape only joints.
	%
	\begin{figure}[t]
		\begin{center}
			\includegraphics[width=0.45\textwidth]{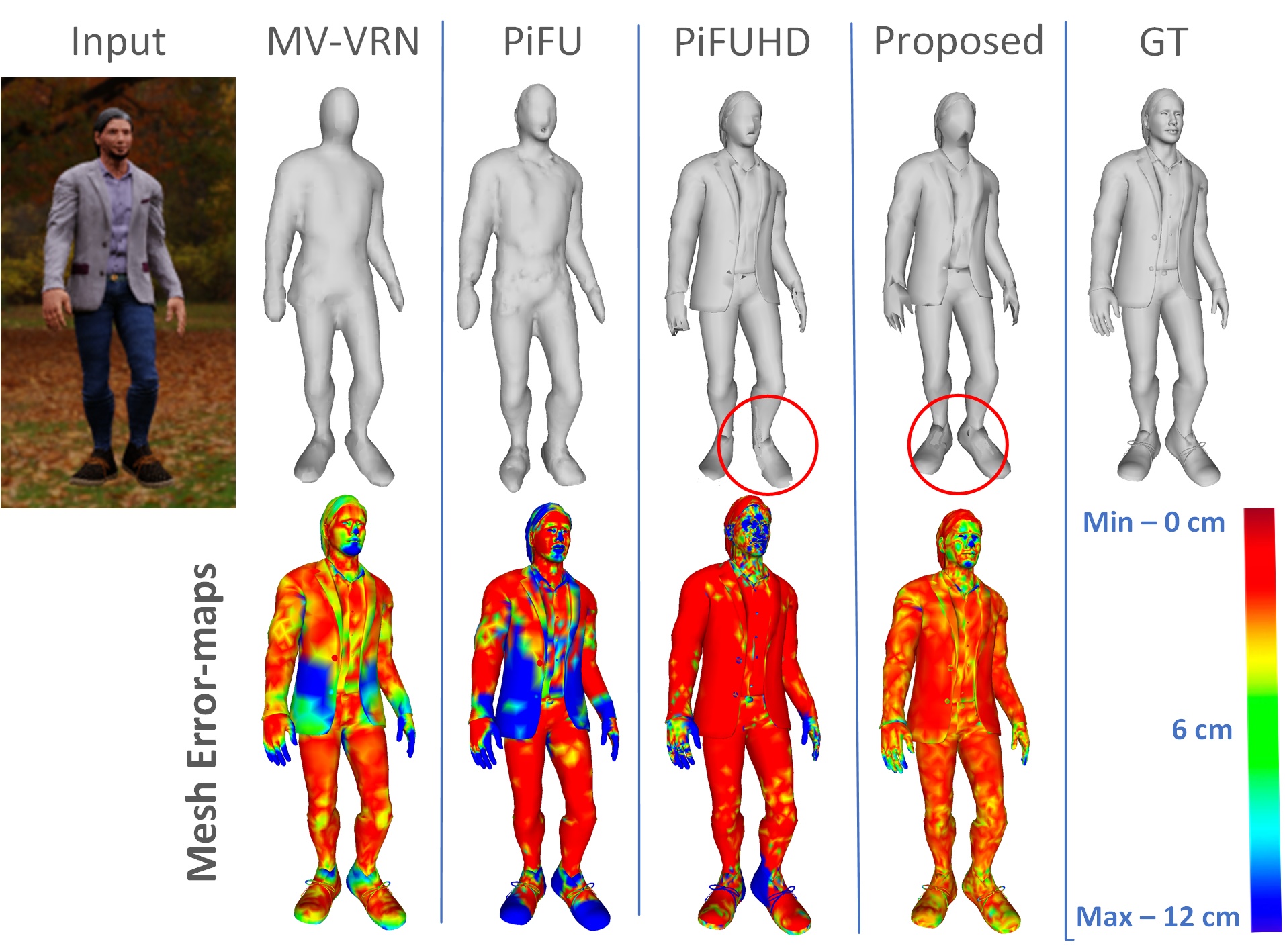}
			\caption{Point to surface (P2S) error maps against ground-truth mesh for proposed method and state-of-the-art methods - GT is ground-truth (error - red - no error and blue - max error ).}
			\label{fig:error}
			\vspace{-0.3cm}
		\end{center}
	\end{figure}
	\begin{figure}[t]
		\begin{center}
			\includegraphics[width=0.48\textwidth]{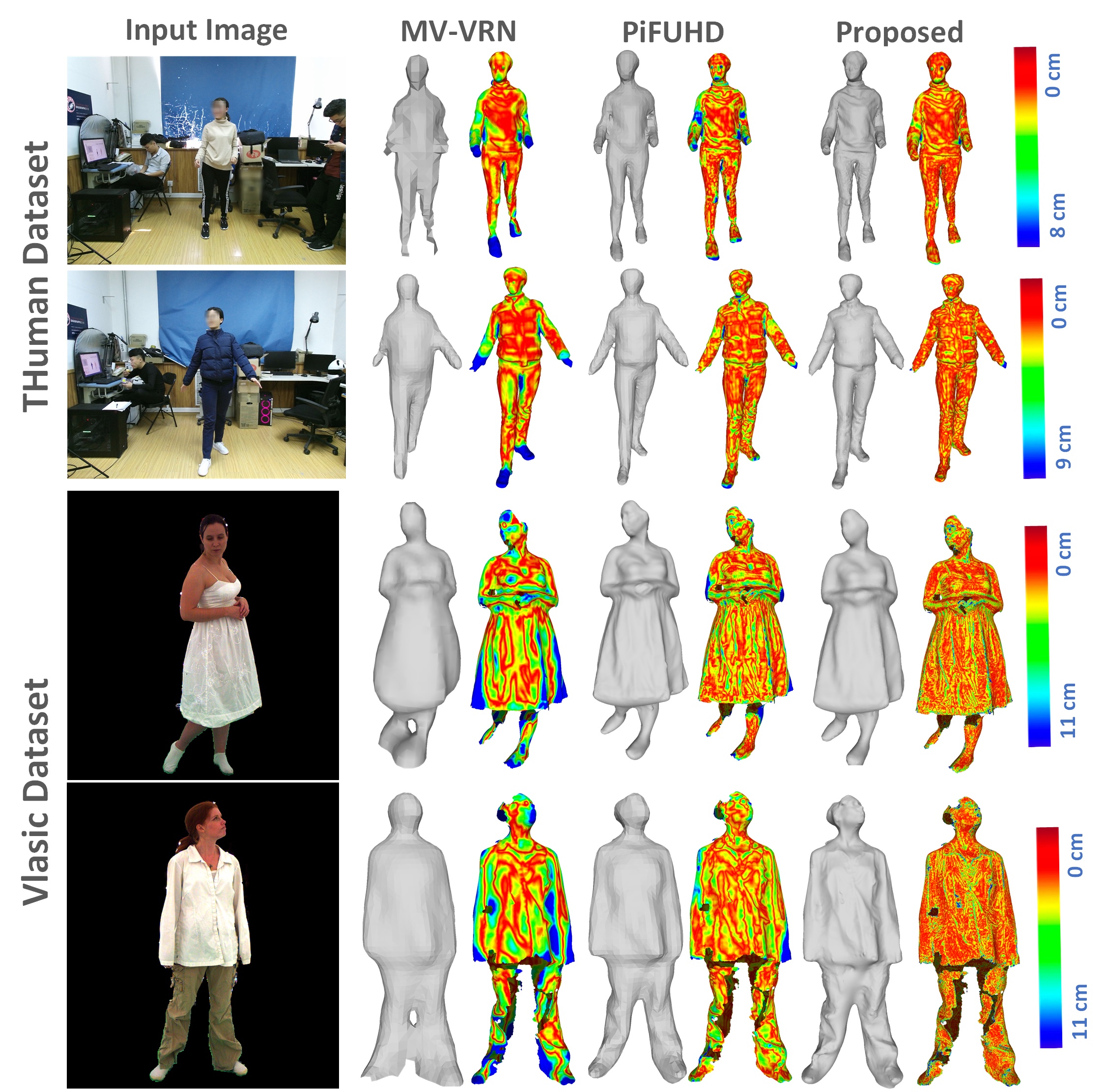}
			\caption{Results and P2S errors against 3D ground-truth for ours, MV-VRN and PiFUHD on real datasets-THuman and Vlasic (error- red - no error and blue - max error)}.
			\label{fig:otherdataset}
			\vspace{-0.6cm}
		\end{center}
	\end{figure}
	\begin{figure*}[t]
		\begin{center}
			\includegraphics[width=0.94\textwidth]{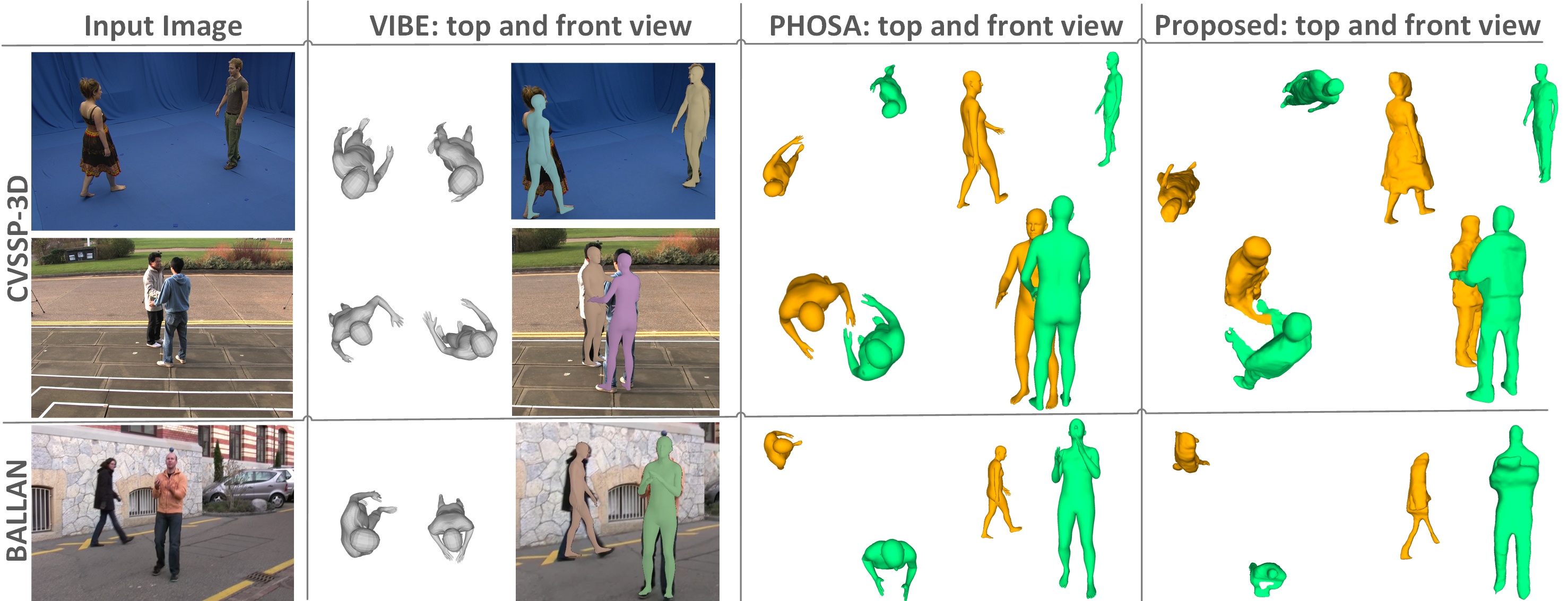}
			\caption{Comparison of proposed multi-human reconstruction against state-of-the-art methods on real datasets - CVSSP3D\cite{Mustafa19} and BALLAN\cite{UnstructuredVBR10}. PHOSA and proposed method give coherent reconstruction. Vibe does not estimate 3D spatial locations of each person.}
			\label{fig:multihuman}
			\vspace{-0.8cm}
		\end{center}
	\end{figure*}
	\subsection{Implementation Details and Architecture}
	\label{sec:implement}
	The proposed network is trained on the novel MPSD dataset, during training the two multitask networks are trained separately. The first multitask segmentation and depth network is given segmentation mask and depth map and the second multitask 3D and pose network is given 3D ground-truth and pose for training. For testing only a single image is passed to the network.
	MPSD is split into training and test sets ($70-30$) such that for each category with $2,3...10$ number of people, $30$ scenes are used to testing and $70$ scenes are used for training. The models in the $30$ test scenes are not seen during training for fair evaluation. The constants are: $\alpha = 0.2, \beta = 0.1, \gamma =0.1, w = 0.001$.
	
	The implicit refinement decoder network consists of stacked linear 1D convolution layers, which takes concatenated features as input and outputs the 3D occupancy. The voxelized 3D estimation network is similar to MV-VRN\cite{caliskan2020multiview}. For further network details including the pose decoder network in pose estimation (Sec. \ref{sec:pose}) and more implementation details please refer to the supplementary material. 
	%
	\begin{figure*}[t]
		\begin{center}
			\includegraphics[width=0.94\textwidth]{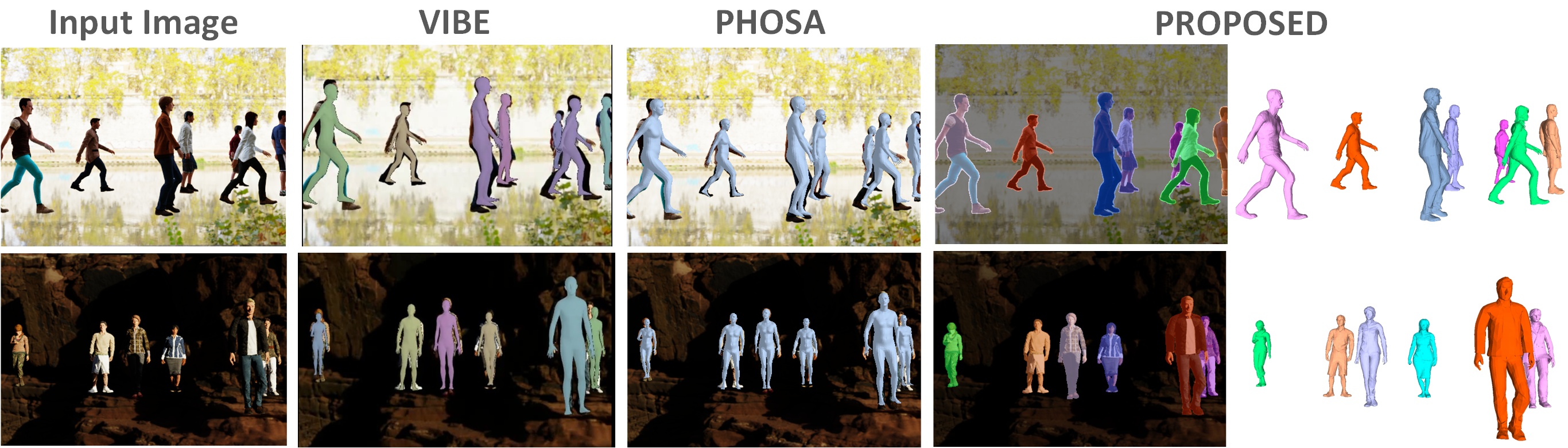}
			\caption{Comparison of mesh projections of single image multi-human reconstruction against the proposed method. Proposed method gives accurate reconstruction with clothing details, unlike PHOSA and Vibe that give SMPL models with no clothing details.}
			\label{fig:mpsdMultiCompare}
			\vspace{-0.8cm}
		\end{center}
	\end{figure*}
	\begin{table}
		\setlength{\tabcolsep}{3pt}
		\scalebox{0.83}{
			\begin{tabular}{l|l|c|l|l|c|l|lll}
				\textbf{Datasets} & \multicolumn{3}{c|}{MPSD} & \multicolumn{3}{c|}{Vlasic} & \multicolumn{3}{c}{THuman} \\ \hline
				\textbf{Method} & CD & 3DIoU & P2S & CD & 3DIoU & P2S & \multicolumn{1}{l|}{CD} & \multicolumn{1}{c|}{3DIoU} & P2S \\ \hline
				MVVRN & 2.11 & 61$\%$ & 2.82 & 2.77 & 55$\%$ & 3.39 & \multicolumn{1}{l|}{2.98} & \multicolumn{1}{c|}{53$\%$} & 3.57 \\ \hline
				PiFUHD & 1.66 & 68$\%$ & 1.93 & 2.14 & 65$\%$ & 2.72 & \multicolumn{1}{l|}{2.56} & \multicolumn{1}{c|}{60$\%$} &  2.97 \\ \hline
				PiFU & 1.98 & 59$\%$ & 2.72 & 2.80 & 48$\%$ & 3.22 & \multicolumn{1}{l|}{3.03} & \multicolumn{1}{c|}{45$\%$} & 3.96 \\ \hline
				DeepH & 3.15 & 48$\%$ & 4.02 & 4.26 & 43$\%$ & 5.58 & \multicolumn{1}{l|}{4.67} & \multicolumn{1}{c|}{39$\%$} & 5.88 \\ \hline
				Ours & \textbf{1.47} & \textbf{71$\%$} & \textbf{1.88} & \textbf{1.93} & \textbf{67$\%$} & \textbf{2.26} & \multicolumn{1}{c|}{\textbf{2.07}} & \multicolumn{1}{c|}{\textbf{65$\%$}} & \textbf{2.86}
			\end{tabular}		
		}
		\caption{Quantitative comparison of our method with state-of-the-art single person methods- CD-Chamfer distance, P2S-Point to surface error, and 3DIoU-3D Intersection of Union. CD and P2S - lower the better, 3DIoU - higher the better. DeepH-DeepHuman}
		\vspace{0.1cm}
		\label{t_quantcomparison}
	\end{table}
	\vspace{-0.3cm}
	\subsection{Comparative Evaluations}
	\noindent
	\textbf{Evaluation on public datasets:} The proposed network is trained on MPSD dataset and fine tuned on the training split of the publically available datasets with 3D, given below: \\
	\textbf{Vlasic \cite{Vlasic_dynamicshape}:} Real single person dataset with 3D\\ 
	\textbf{THuman\cite{zheng2019deephuman}:} Real single person dataset with 3D \\
	\textbf{CVSSP3D \cite{Mustafa19}:} Real multiple people dataset with 3D \\
	\textbf{Ballan \cite{UnstructuredVBR10}:} Real multiple people dataset with 3D state-of-the-art single person shape estimation methods.\\
	\textbf{Single person evaluation on MPSD dataset:} We crop the image of one person with no occlusions from proposed the MPSD dataset for fair single person reconstruction evaluation with existing state-of-the-art methods - PiFU \cite{Saito_2019_ICCV}, PiFUHD \cite{saito2020pifuhd}, MV-VRN \cite{caliskan2020multiview}, and DeepHuman\cite{zheng2019deephuman}. PiFU and MV-VRN are trained on the MPSD datasets and due to unavailability of code DeepHuman and PiFUHD are only tested on MPSD. Three error metrics are computed using the ground-truth 3D models to measure the quality of shape reconstruction: Chamfer Distance (CD), Point to surface errors (P2S) \cite{saito2020pifuhd} and 3D Intersection of Union (3D IoU) \cite{jatavallabhula2019kaolin}. Qualitative comparison is shown in Fig. \ref{fig:comparesingle} and quantitative evaluation is shown in Fig. \ref{fig:error} and Table \ref{t_quantcomparison}. The comparison demonstrates that the proposed method gives a high-resolution reconstruction comparable to PiFUHD and significantly better than PiFU, MV-VRN, DeepHuman. The quality and completeness of the reconstruction in the unseen parts of the object are better compared to PiFUHD (side-view in Fig. \ref{fig:comparesingle}). Results and comparative evaluation on Vlasic and THuman datasets against PiFUHD and MV-VRN in Fig. \ref{fig:otherdataset} shows that the proposed method significantly outperforms existing methods on real datasets. \\
	\textbf{Multi person evaluation on MPSD dataset:} The results against multiple people 3D shape estimation methods PHOSA\cite{Zhang2020Perceiving3H} and VIBE\cite{Kocabas2020VIBEVI} are shown in Fig. \ref{fig:multihuman}. Quantitative comparison is given in Table \ref{t_quantcomparemulti} against Chamfer distance, P2S and 2D Intersection of Union \cite{Mustafa19}. Both PHOSA and VIBE estimate the SMPL model of each person in the scene unlike the proposed method which gives model-free realistic reconstruction of people, which align with the object boundaries more closely especially in the case of loose clothing, as seen in Fig. \ref{fig:mpsdMultiCompare}. Results on real datasets CVSSP3D and Ballan datasets against PHOSA and VIBE are illustrated in Fig. \ref{fig:multihuman} and on MPSD dataset is shown in Fig.\ref{fig:mpsdMultiCompare}, demonstrating that the proposed method outperforms existing methods on both real and synthetic datasets in quality and coherency of the reconstruction. We also compare the 6DOF spatial location and orientation estimated from the proposed method and PHOSA in Table \ref{t_pose} using AUC metric defined in \cite{Wang2019DenseFusion6O} for synthetic and real dataset and the results are comparable.
	%
	\begin{table}[h]
		\setlength{\tabcolsep}{3pt}
		\scalebox{0.82}{
			\begin{tabular}{l|l|c|l|l|c|l|lll}
				\textbf{Datasets} & \multicolumn{3}{c|}{MPSD} & \multicolumn{3}{c|}{CVSSP3D} & \multicolumn{3}{c}{Ballan} \\ \hline
				\textbf{Method} & CD & 2DIoU & P2S & CD & 2DIoU & P2S & \multicolumn{1}{l|}{CD} & \multicolumn{1}{l|}{2DIoU} & P2S \\ \hline
				PHOSA & 5.77 & 73$\%$ & 6.92 & 7.73 & 69$\%$ & 8.32 & \multicolumn{1}{l|}{9.11} & \multicolumn{1}{c|}{67$\%$} & 10.04 \\ \hline
				Vibe & 6.12 & 68$\%$ & 7.44 & 8.20 & 63$\%$ & 9.14 & \multicolumn{1}{l|}{10.36} & \multicolumn{1}{c|}{64$\%$} & 11.38 \\ \hline
				Ours & \textbf{1.47} & \textbf{84$\%$} & \textbf{1.88} & \textbf{1.97} & \textbf{80$\%$} & \textbf{2.31} & \multicolumn{1}{l|}{\textbf{2.27}} & \multicolumn{1}{c|}{\textbf{78$\%$}} & \textbf{3.02}
			\end{tabular}
		}
		\caption{Comparison of the proposed method against multiple people reconstruction methods- CD is Chamfer distance $\downarrow$, 2DIoU is 2D Intersection of Union $\uparrow$, and P2S is Point to surface error $\downarrow$.}
		\vspace{0.2cm}
		\label{t_quantcomparemulti}
	\end{table}
	\begin{table}[h]
		\setlength{\tabcolsep}{4pt}
		\scalebox{0.9}{
			\begin{tabular}{l|c|c|l|c|c}
				& MPSD & CVSSP3D &  & MPSD & CVSSP3D \\ \hline
				PHOSA & 75.6 & 80.4 & Proposed & 79.8 & 82.1
			\end{tabular}
		}
		\caption{Comparison of the spatial location and orientation of the proposed method against PHOSA for the AUC metric\cite{Wang2019DenseFusion6O}.}
		\vspace{0.2cm}
		\label{t_pose}
	\end{table}
	\begin{table}[h]
		\setlength{\tabcolsep}{4pt}
		\scalebox{0.93}{
			\begin{tabular}{l|l|l|c|c}
				& CD $\downarrow$ & P2S $\downarrow$ & 3DIoU $\uparrow$ & 2DIoU $\uparrow$ \\ \hline
				Proposed & \textbf{1.47} & \textbf{1.88} & \textbf{71$\%$} & \textbf{84$\%$} \\ \hline
				W/o $L_{OS}$ \&  implicit & 2.11 & 2.82 & 61$\%$ & 72$\%$ \\ \hline
				W/o Implicit & 1.98 & 2.71 & 62$\%$ & 74$\%$\\ \hline
				W/o $L_{OS}$ & 1.63 & 2.44 & 65$\%$ & 79$\%$ \\ \hline
				Implicit with 3D \& Depth & 1.71 & 2.42 & 64$\%$ & 79$\%$ \\ \hline
				Implicit with 3D \& RGB  & 1.65 & 2.27 & 67$\%$ & 80$\%$ \\ \hline
				If-net \cite{Chibane_2020_CVPR} & 1.78 & 2.59 & 63$\%$ & 77$\%$ 
		\end{tabular}}
		\caption{Ablation study of the proposed method on MPSD dataset.}
		\vspace{-0.2cm}
		\label{t_ablation}
	\end{table}
	\subsection{Ablation Study}
	\label{sec:ablation}
\begin{figure}[h]
		\begin{center}
			\includegraphics[width=0.48\textwidth]{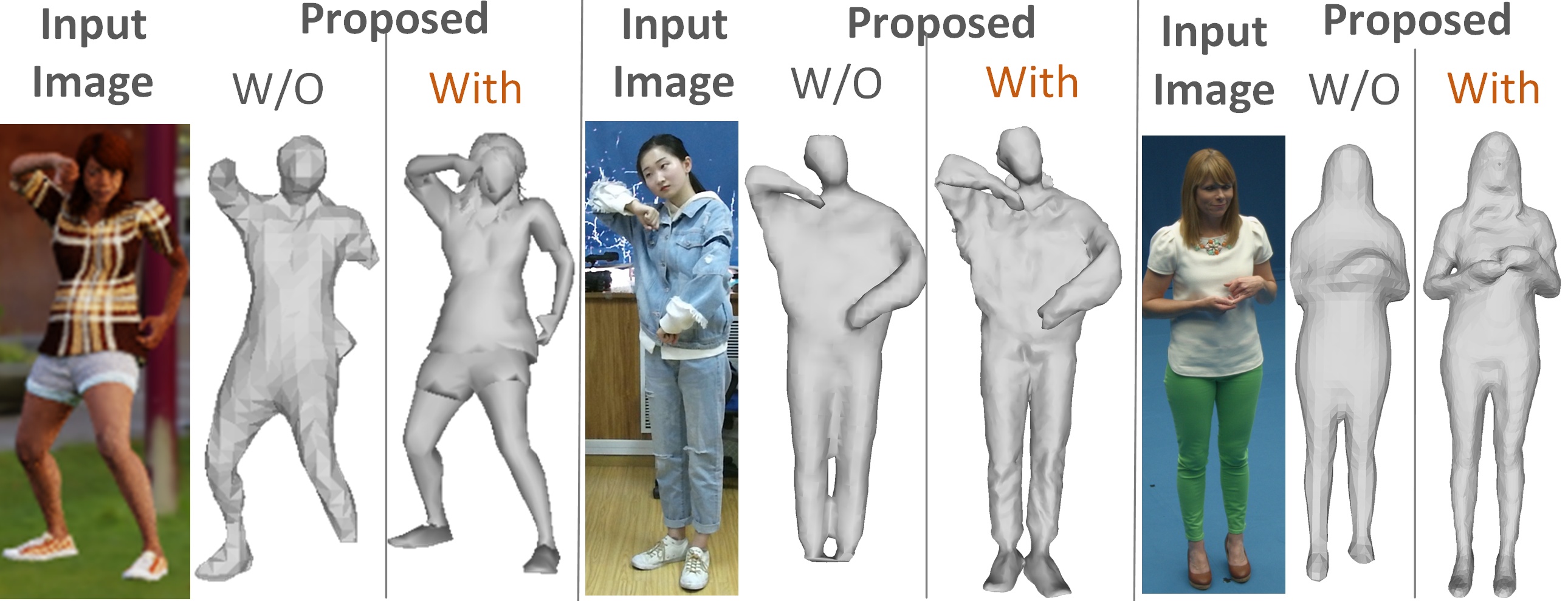}
			\caption{Results of proposed method with and without implicit refinement on synthetic MPSD (left) and real THuman and CVSSP3D (right) datasets.}
			\label{fig:implicitablation}
			\vspace{-0.2cm}
		\end{center}
	\end{figure}

	The abaltion for implicit 3D reconstruction (Sec. \ref{sec:3d}) of the proposed method with and without the implicit refinement is shown in Fig. \ref{fig:implicitablation}. There is a considerable improvement in the details, completeness and quality of the surface reconstruction. In the implicit refinement encoded features are input from RGB image, depth and 3D voxels. We present ablation without the silhouette loss in the voxelized 3D estimation ($L_{OS}$), without implicit refinement, with implicit refinement only with 3D and depth features  (Implicit-3D \& Depth), with implicit refinement with 3D and RGB features (Implicit-3D \& RGB) and the proposed method in Table \ref{t_ablation} for all error metrics defined before - CD, P2S, 3DIoU and 2DIoU. The proposed methods performs the best in terms of quality and completeness of the 3D shape when all 3 features are encoded in the implicit refinement network (Fig. \ref{fig:ImplicitR}).
	%
	\begin{figure}[h]
		\begin{center}
			\includegraphics[width=0.48\textwidth]{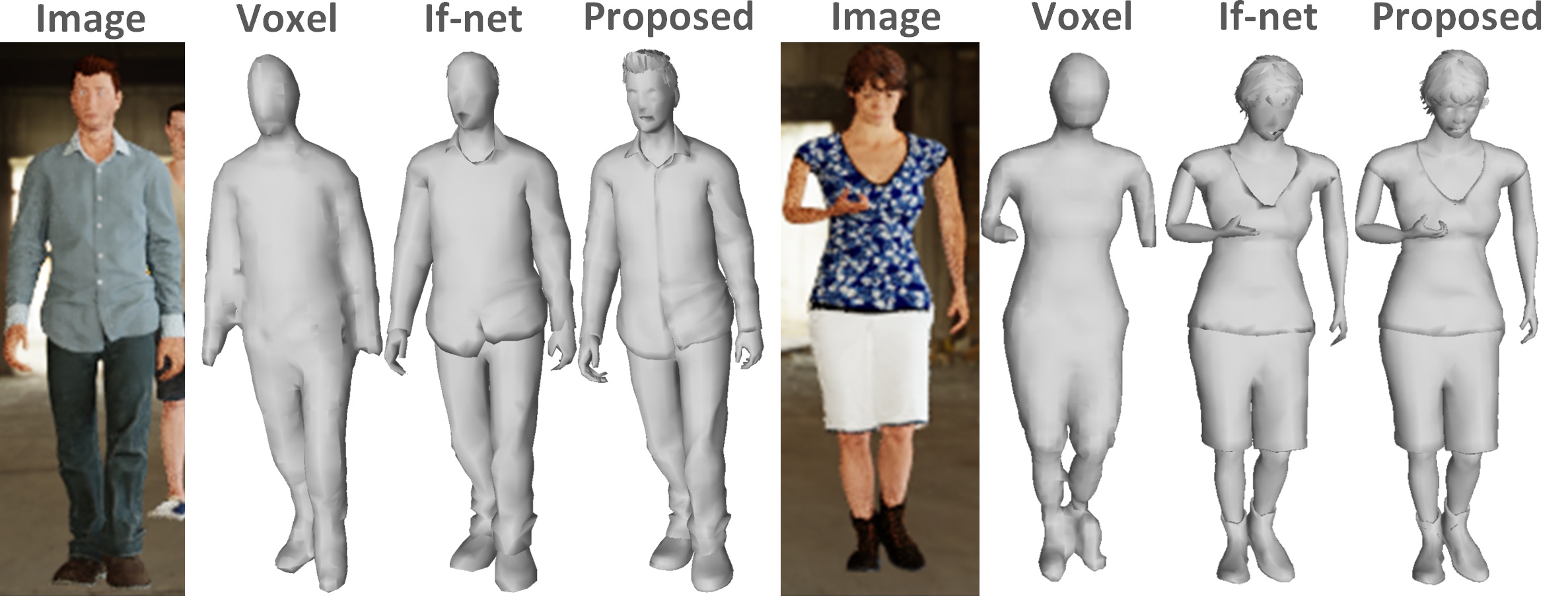}
			\caption{Comparison of proposed approach with If-net \cite{Chibane_2020_CVPR}}
			\label{fig:ifnetcompare}
			\vspace{-0.2cm}
		\end{center}
	\end{figure}
	
	We also compare our implicit refinement with If-net \cite{Chibane_2020_CVPR} in Fig. \ref{fig:ifnetcompare}. The input to the implicit refinement network and If-net is the voxelized output from the first stage of the implicit 3D reconstruction (Sec. \ref{sec:3d}). The 3D surface from the proposed method has more details compared to If-net.

	\noindent
	\textbf{Limitations:}
	Like existing methods, the proposed method cannot handle extreme poses (handstand) and heavy occlusions ($>50\%$). As it's a multi-level approach like existing methods \cite{Saito_2019_ICCV}, it relies on the success of instance segmentation and depth. Also no temporal information is exploited potentially leading to a temporally incoherent output.
	%
	\section{Conclusion}
	This paper presents an end-to-end learning framework for spatially coherent model-free implicit reconstruction of a multi-person single image. The method gives realistic implicit reconstructions of people with loose clothing and hair without any manual intervention compared to previous SMPL based approaches and works with partially occluded people. The proposed methodology combines the advantages of explicit volumetric representations to reconstruct a wide range of poses and implicit function representations for reconstruction of shape detail. A multitask network is used to simultaneously estimate 3D shape and 6DOF spatial location and orientation of each person in the image for a spatially coherent multi-human reconstruction. The proposed network is trained on our novel \textit{MPSD} dataset with image-3D ground-truth pairs.
	High-resolution, robust and spatially coherent reconstructions are demonstrated on synthetic and real datasets on complex multi-person single images with partial occlusions, in a variety of clothing, poses, and scenes.
	Extensive comparative evaluation with single person 3D shape estimation methods demonstrates improvement in the accuracy, completeness and detail of the reconstruction. Comparative evaluation with state-of-the-art multi-human reconstruction methods shows that the proposed approach achieves a significantly better reconstruction of clothed humans as the existing methods only give SMPL model reconstruction of each person in the scene.
	
	\noindent
	{\bf Acknowledgement:} Supported by Dr Mustafa's RAEng Fellowship and EPSRC Platform Grant EP/P022529/1.
	%
	
	{\small
		\bibliographystyle{ieee_fullname}
		\bibliography{egbib}
	}
	
\end{document}